\documentclass[twoside,twocolumn]{article}

\usepackage{blindtext} % Package to generate dummy text throughout this template 

\usepackage[scaled=.95]{newtxtext,newtxmath} % Font is Times New Roman equivalent, scaled for font size
\usepackage[T1]{fontenc} % Use 8-bit encoding that has 256 glyphs
\usepackage{microtype} % Modify fonts

\usepackage[polish,slovene,english]{babel} % Language hyphenation and typographical rules
\usepackage[utf8]{inputenc} % Also for other symbols

\usepackage{graphicx} % Inserting graphics
\usepackage{makecell} % Table cells
\usepackage{tabularx} % Table formatting
\usepackage{booktabs}
\usepackage{bchart} % Barchart
\usepackage{multirow} % Table multirow headings

\usepackage[hang]{footmisc} % Footnote indentation
\usepackage[hmarginratio=1:1,margin={1.9cm,1.9cm},top=26mm,bottom=26mm,columnsep=20pt]{geometry}
\usepackage[hang, small,labelfont=bf,up,textfont=bf,up,format=plain]{caption} % Custom captions under/above floats in tables or figures
\usepackage{booktabs} % Horizontal rules in tables

\usepackage{lettrine} % The lettrine is the first enlarged letter at the beginning of the text

\usepackage{enumitem} % Customized lists
\setlist[itemize]{noitemsep} % Make itemize lists more compact

\usepackage{abstract} % Allows abstract customization
\usepackage{titlesec} % Allows customization of titles
\titleformat{\section}[block]{\large\bfseries}{\thesection}{1em}{\MakeUppercase}{} % Change the look of the section titles
\titlespacing*\section{0pt}{6pt}{5pt}
\titleformat{\subsection}[block]{\large\bfseries}{\thesubsection}{1em}{} % Change the look of the section titles
\titlespacing*\subsection{0pt}{6pt}{5pt}
\titleformat{\subsubsection}[runin]{\normalsize\itshape}{\thesubsubsection}{1em}{} \titlespacing*\subsubsection{0pt}{6pt}{5pt}

\usepackage{fancyhdr} % Headers and footers
\fancypagestyle{specialfooter}{%
  \fancyhf{}
  
  \fancyfoot[L]{©2020 Copyright held by the authors. This is the author's version of the work. It is posted here for your personal use. Not for redistribution. The definitive version was published in the Proceedings of the 29th ACM International Conference on Information and Knowledge Management, \url{https://doi.org/10.1145/3340531.3412783}.}
}

\usepackage{titling} % Customizing the title section

\usepackage{hyperref} % For hyperlinks in the PDF

\newlength{\bibitemsep}
\newlength{\bibparskip}
\let\oldthebibliography\thebibliography
\renewcommand\thebibliography[1]{%
  \oldthebibliography{#1}%
  \setlength{\parskip}{\bibitemsep}%
  \setlength{\itemsep}{\bibparskip}%
}

\title{MLM: A Benchmark Dataset for Multitask Learning with Multiple Languages and Modalities} % Article title
\author{%
\textsc{Jason Armitage}\thanks{Denotes equal contribution to this research} \\
\normalsize University of Bonn \\
\normalsize Germany
\and 
\textsc{Endri Kacupaj}\footnotemark[1]\ \\
\normalsize University of Bonn \\
\normalsize Germany
\and 
\textsc{Golsa Tahmasebzadeh}\footnotemark[1]\ \\
\normalsize TIB -- Leibniz Information \\ \normalsize Center for Science and \\ \normalsize Technology \\
\normalsize Germany
\and 
\textsc{Swati}\ \\
\normalsize Jožef Stefan Institute \\
\normalsize Slovenia 
\and 
\textsc{Maria Maleshkova}\ \\
\normalsize University of Bonn \\
\normalsize Germany 
\and
\textsc{Ralph Ewerth}\ \\
\normalsize TIB -- Leibniz Information \\ \normalsize Center for Science and \\ \normalsize Technology \\
\normalsize Germany 
\and
\textsc{Jens Lehmann}\ \\
\normalsize University of Bonn \\
\normalsize Fraunhofer IAIS \\
\normalsize Germany
}
\date{} % Leave empty to omit a date

\begin{document}

% Print the title
\maketitle
\pagestyle{empty} % Remove page numbers
\thispagestyle{specialfooter}

\begin{abstract}
\vspace*{-.9em}
\noindent 
In this paper, we introduce the MLM (Multiple Languages and Modalities) dataset - a new resource to train and evaluate multitask systems on samples in multiple modalities and three languages. The generation process and inclusion of semantic data provide a resource that further tests the ability for multitask systems to learn relationships between entities. The dataset is designed for researchers and developers who build applications that perform multiple tasks on data encountered on the web and in digital archives. A second version of MLM provides a geo-representative subset of the data with weighted samples for countries of the European Union. We demonstrate the value of the resource in developing novel applications in the digital humanities with a motivating use case and specify a benchmark set of tasks to retrieve modalities and locate entities in the dataset. Evaluation of baseline multitask and single task systems on the full and geo-representative versions of MLM demonstrate the challenges of generalising on diverse data. In addition to the digital humanities, we expect the resource to contribute to research in multimodal representation learning, location estimation, and scene understanding.
\end{abstract}

% Sections
% Section

\section{Introduction}

At the core of recent research in machine learning is the proposal that systems trained on large sets of data will display forms of the human ability to generalise. Learning algorithms will enable systems to infer abstract representations from raw data and tackle a wide range of tasks~\cite{bengio_learning_2009}. The web provides an abundance of data for training models that enable functions in application pipelines. Data availability then is no impediment to the aims of machine learning in relation to these applications - but there remain two important ways in which systems fail to generalise. 

The first of these is the inability for systems to cope with diversity in real-world data~\cite{becker_new_2016,xue-wen_chen_big_2014}. Even small changes in the distributions of data lead to steep declines in the performance of machine learning models. Information on the web and maintained in archives is present in a diverse range of modalities including language, images, and numerical data. Variety is also observed in each modality taken in isolation. Consider text in documents stored and studied by cultural and academic institutions. A system that is designed to predict contextual information - in this case, locations referenced in the documents~\cite{alex2015adapting} - may encounter text written in several languages and writing systems. Additional inputs with relevant information can be stored in databases or as structured text in a knowledge base. Machine learning systems that are robust to disparities between and within modalities can make use of a higher proportion of the information present across a range of materials~\cite{baltruvsaitis2018multimodal}. 

\vspace{-1.5pt}

The second way in which current systems fail to generalise is in using representations learned during training to perform new tasks. Analysis of machine learning in application workflows reveals a host of disparate models - each trained on homogeneous data to conduct specialised tasks~\cite{prado_m}. In this approach, learning is siloed with the result that computation and data required for training and inference multiply - and there is limited sharing of parameters and representations across pipelines. Systems that conduct multiple tasks hold the promise of simplifying application architectures and improving efficiency in the use of resources. In the case of researchers conducting cultural and historical analysis, this also presents the advantages of studying or accessing materials in multiple formats~\cite{pauwels2012multimodal} such as textual documents and images that reference the same subject.

The Multiple Languages and Modalities (MLM) dataset consists of text in three languages (English, French, and German), images, location data, and knowledge graph triple classes (ie component entities of triples for the sample entity that are stored in Wikidata). The resource is designed to evaluate multitask systems in relation to single task alternatives when trained simultaneously on sequences of tasks - in this case, a system that performs both cross-modal retrieval and location estimation. Multitask learning aims to benefit from commonalities in input data~\cite{caruana1997multitask} and MLM is derived from knowledge graph entities that connect all of the samples in the data. Relationships between entities are further manifested in triple classes. 

\subsection{Contributions}

We identify four contributions resulting from this research.
\begin{itemize}[topsep=0pt]
    \item The first contribution is the generation process, which provides a consistent and reproducible approach to build a resource of diverse data with semantic relationships. As detailed below, this process will accelerate the release of future versions of MLM. 
    \item The second contribution is the release of a novel dataset with a range of modalities and languages that - to our knowledge - is unique in the research literature. The resource tests the ability for multitask systems to exploit both formal and indirect relationships between samples. 
    \item The third contribution is the specification of a benchmark series of tasks to evaluate multitask systems that characterise applications used and built by researchers and developers in the digital humanities. The resource presents a unique option to this community to develop novel applications that make use of multimodal frameworks for the assessment of cultural and historical phenomena \cite{pauwels2012multimodal}. 
    \item The fourth contribution is building a multitask system that combines several methods to perform in concurrence cross-modal retrieval and location estimation on highly diverse data. 
\end{itemize}

\subsection{Use Case}
We present a use case that applies to the benchmark evaluation specified in this research and characterises the novel applications enabled by a multitask approach to diverse data in a real-world scenario. The MLM resource and benchmark tasks also extends to use cases in several other areas highlighted in Section \ref{impact}.

Relations between media and geospatial location underpin applications and projects developed for cultural and historical analysis. The Edinburgh Geoparser~\cite{alex2015adapting} and Perdido ~\cite{moncla2017automated} are recent instances of applications that retrieve locations from multilingual literary texts sourced in books and journals. Identifying locations depicted in visual materials is also a core function in applications that georeference historical images on landscapes and urban areas~\cite{blanc2018semi,harrach2019interactive}. In the multitask system framework presented below (see Figure \ref{fig:mlm_generation}), we enable geoparsing and georeferencing in two novel ways. First the researcher has the option of starting a query either with a text or an image. During the retrieval task (see Figure \ref{fig:mlm_process}), related materials in the missing modality are returned. This creates the opportunity to conduct a multimodal analysis on the entity of interest~\cite{pauwels2012multimodal}. Second a multitask framework returns a predicted location for both inputs. 

\vspace{0.4cm}
\begin{center}
\captionsetup{type=figure}
\includegraphics[width=0.35\textwidth,height=0.25\textwidth]{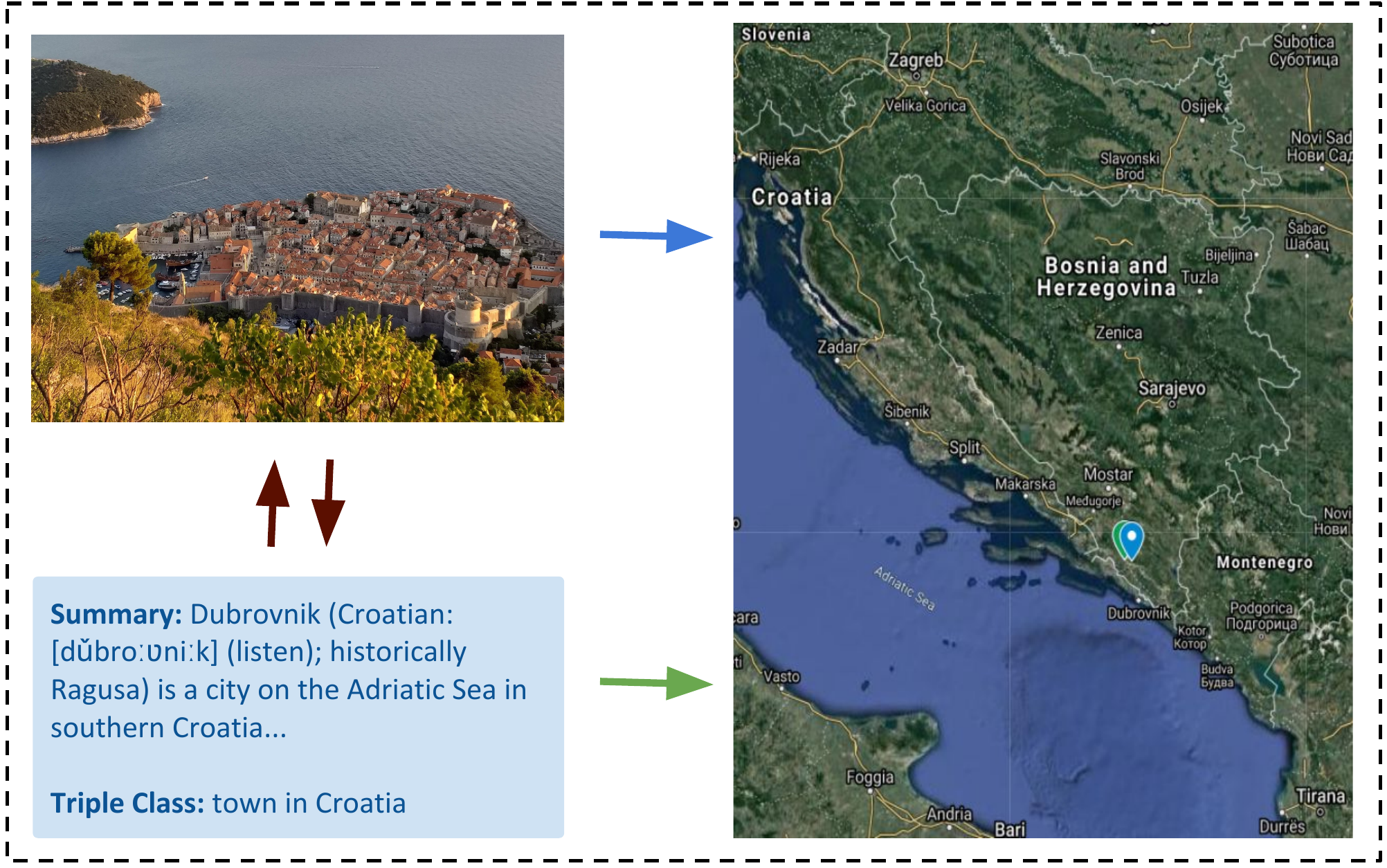}
\caption{Cross-modal retrieval and location estimation performed by the Multitask IR+LE framework.} 
\label{fig:mlm_process}
\end{center}

% Section

\section{Impact} \label{impact}
In this section, we highlight research areas where the MLM resource enables innovation and assess the benefits of the dataset in relation to existing resources for evaluating multitask systems.

We have outlined above new approaches in the digital humanities that are enabled by the dataset and benchmark evaluation tasks. Multitask learning systems that learn on multimodal data are also an active area of research in relation to multimodal representation learning, location estimation, and scene understanding~\cite{baltruvsaitis2018multimodal, kuga2017multi}. MLM is further designed to evaluate the ability for multitask systems to leverage relationships between constituent entities in data and knowledge graph properties used in the generation process. Multitask learning systems that exploit these relationships as a signal are positioned to deliver additional benefits to applications that rely on semantic data and knowledge graphs, which include recommender systems, mobile information retrieval, and bioinformatics~\cite{ali2019keen}. 

Resources for training and evaluating multitask learning systems are characterised by homogeneity both in modalities and languages. The Office-Home and Computer Survey are examples of resources to train text-only systems~\cite{zhang_survey_2018}. Similarly multitask systems for computer vision turn to benchmark datasets where images are the primary modality~\cite{caputo2014imageclef, long2017learning}. Multimodal datasets employed for multitask learning consist predominantly of two modalities~\cite{li2018vqa,choi_intra-modal_2019} and a single language. Although ImageCLEF runs challenges (see Table \ref{tab:mtl_datasets}) with textual samples in multiple languages, these instances also focus on bimodal learning~\cite{tsikrika2011overview}. The Multiple Language and Modalities dataset, which we propose, approximates the diversity of online data and those stored in digital archives or databases in presenting samples in three modalities with multilingual text and semantic data. To our knowledge, this resource is unique in presenting an opportunity to train multitask systems on visual, textual, spatial and semantic web inputs for multimodal, and multilingual learning.

Geographical location is central to both the resource and task presented in this research. We are motivated by on-going investigations on representation bias in machine learning datasets to present a second geo-representative version of our resource. Representation bias has been found to penalise minority classes in classification tasks \cite{suresh2019framework}. MLM-irle-gr is a subset of the full data with balanced sample sizes of entities for the 28 countries in the European Union. This enables governmental and commercial organisations that provide applications and services in this region to minimise the impact of representation bias resulting from training data when evaluating multitask learning systems.

\begin{table}[]
\caption{Multimodal and Multilingual Datasets}
%\vspace{-1em}
\label{tab:mtl_datasets}
\begin{tabularx}{\columnwidth}{lccc}
\toprule
\textbf{Resource} & \textbf{Data Types} & \textbf{Language} & \textbf{Samples} \\ \hline
\textbf{MLM} & \makecell{\begin{tabular}[c]{@{}c@{}}Text, Images,\\ Geocoordinates\\Triple Classes\end{tabular}} & DE, EN, FR & 236k \\
\textbf{ImageCLEF} & Text, Images & DE, EN, FR & 237k \\
\textbf{IKEA} & Text, Images & DE, EN, FR & 4k \\
\bottomrule
\end{tabularx}
\vspace{-4mm}
\end{table}

% Section

\section{MLM - Multiple Languages and Modalities}
In this research we propose a benchmark to evaluate the strengths of multitask systems to generalise on diverse data. This section starts by defining the tasks constituting the evaluation, continues with an introduction of the dataset, and concludes with details on the generation process.

\subsection{Benchmark Evaluation}
The benchmark evaluation is composed of two tasks that are performed in conjunction to retrieve and locate entities represented by multimodal and multilingual inputs. The first task consists of cross-modal retrieval where a representation in an input modality $x_i$ returns the target $y_i$ in a corresponding modality. In this case, the inputs are one from a pair $P=\{(x^u_i, x^t_i)\}^n_i=1$, where $x^u_i$ is an image and $x^t_i$ is a combination of structured $x^r_i$ and unstructured $x^s_i$ textual inputs. The objective is to return the corresponding target $y_i$ at the lowest position in the ranking of returned predictions. The objective of the second task is to use the same pair of inputs $P=\{(x^u_i, x^t_i)\}^n_i=1$ to perform a classification on a set of geo-cells $y=(y_1,y_2,...y_n)$ - each of which groups a set of entities based on location given as geocoordinates. The final objective of the evaluation is to return results that surpass the baseline system presented below on both tasks. Additional details of the constituent tasks are included in Section \ref{taskssec}. 

\subsection{Dataset}
The Multiple Languages and Modalities comprises data points on 236k human settlements for evaluating and optimising multitask learning systems. MLM presents a dataset with a high level of diversity in terms of modality and language. For each entity, we have extracted text summaries, images, coordinates, and their respective triple classes. Text summaries are available in three languages (English, French, and German) with each entity having between one and three language entries. 

\begin{figure}[!ht]
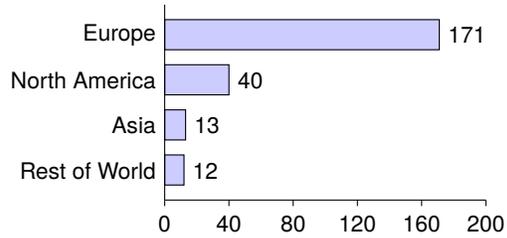

\begin{bchart}[step=40,max=200,width=0.3\textwidth,scale=0.8]
    \bcbar[label=Europe]{171}
        \smallskip
    \bcbar[label=North America]{40}
        \smallskip
    \bcbar[label=Asia]{13}
        \smallskip
    \bcbar[label=Rest of World]{12}
%\bcxlabel{MLM: Number of Entities by Continent (000).} 
\end{bchart}
\vspace{-1em}
\caption{MLM: Number of Entities by Continent (000).} 
\label{fig:mlm_cont}
\vspace{-4mm}
\end{figure}

Human settlements from all continents are provided in the full dataset with 72\% located in Europe (see Figure \ref{fig:mlm_cont}). MLM-irle is a version of the dataset generated for the benchmark evaluation tasks and features geo-cells required for location estimation. To serve organisations that focus on the European Union, we have also created a version of the dataset - MLM-irle-gr (ie geo-representative) - that provides a geographically balanced coverage of human settlements in this region. MLM-irle-gr contains data on 24k human settlements across the EU weighted in relation to the population count\footnote{\url{https://population.un.org/wpp/Download/Standard/Population/}} for each of the 28 countries.
\begin{table}[]
\caption{MLM - Dataset Details}
\label{tab:mlm_dataset}
\begin{tabular}{lccc}
\toprule
\textbf{Num. of} & \textbf{MLM} & \textbf{MLM-irle} & \textbf{MLM-irle-gr} \\ \hline
\textbf{Entities} & 236496 & 218681 & 22501 \\
\textbf{Images} & 412422 & 314533 & 31621 \\
\textbf{Summaries} & 497899 & 462328 & 47508 \\
\textbf{Triple classes} & 1685 & 1655 & 452 \\
\bottomrule
\end{tabular}
\label{table:mlm_datasets}
\vspace{-4mm}
\end{table}

\subsection{Generation Process}
In this first release, a core aim of the research was to develop a generation process that enables fast and consistent scaling for future versions. The process outlined also results from a core requirement for the resource: to evaluate the ability of multitask systems to incorporate information on entities with semantic relations. In this section, we detail stages in the framework for dataset generation. Our selection of the Wikidata Knowledge Graph\footnote{\url{https://www.wikidata.org/}} as the primary source was conditioned on the requirement for data with semantic relations between entities and the ability enabled by SPARQL to target data that met our research aims.
\begin{center}
\captionsetup{type=figure}
\includegraphics[width=0.22\textwidth,height=0.29\textwidth]{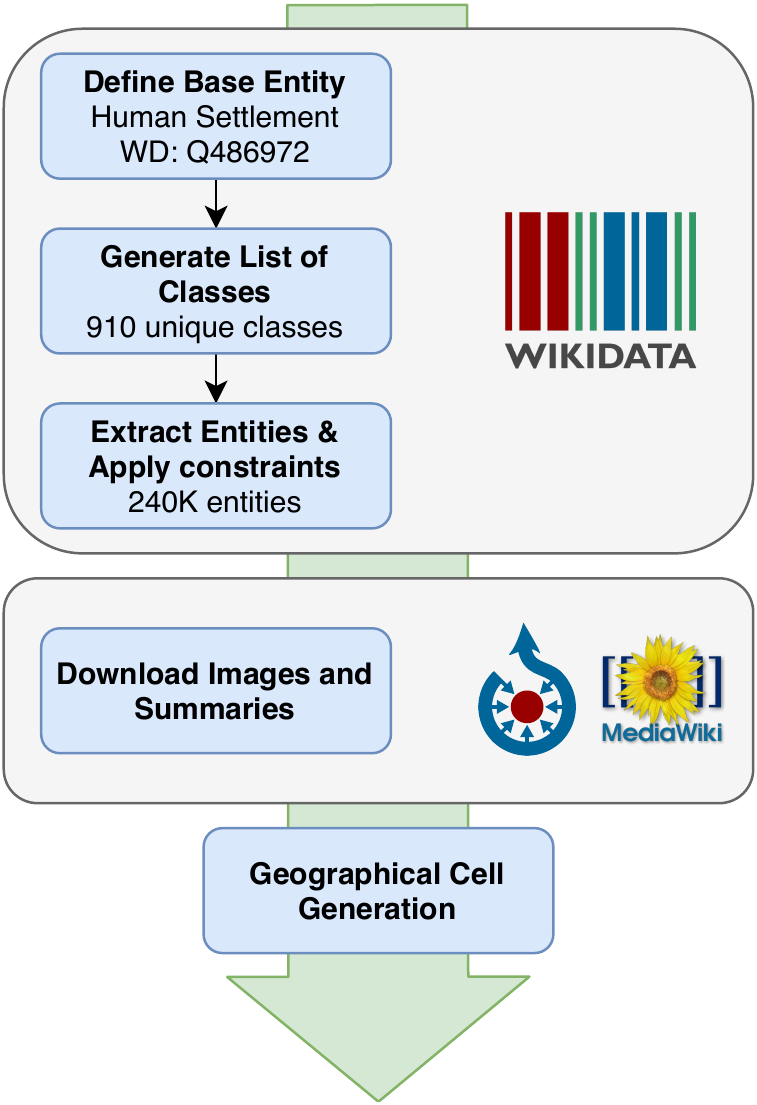}
\caption{MLM dataset generation process.} 
\label{fig:mlm_generation}
\end{center}
\subsubsection{Define base entity type.}
The first step consisted of defining the base entity type that best supported developing applications that make use of geographical location. We selected the Wikidata entity human settlement (ID Q486972)\footnote{IDs are Wikidata identifiers \url{https://www.wikidata.org/wiki/Wikidata:Identifiers}} based on the number of entities returned relative to other candidate base entity types. In the second step, instances of human settlement base type with targeted modalities were extracted resulting in over half a million entities. We applied constraints to extract entities with a Wikipedia article in at least one target language (EN, DE, FR), a minimum of one image, and a set of geocoordinates resulting in 40k human settlements. 

\subsubsection{Generate a list of classes.}
We increased the scope and granularity of human settlements by extracting entity types that are subclass of the the base type. Entities are returned for triples upto three hops from the starting entity. An example of additions is the commune of Italy (ID Q747074) - a subclass of the entity municipality (ID Q15284), which is a subclass of the human settlement base type. This stage resulted in a set of 910 unique classes where all entities derive from the base type by either one, two, or three hops.

\subsubsection{Extract entities.}
In the extraction stage, we run queries on all entities that are an instance of at least one of the 910 unique classes in the generated list. This returned a set of 2.5M entities. A final application of the constraints related to targeted data types generated a collection of data that constituted 240k entities.

\subsubsection{Download images and summaries.}
The final stage consists of downloading images and text summaries for the extracted entities. While querying Wikidata we were able to also extract WikiMedia Commons\footnote{\url{https://commons.wikimedia.org/wiki/Main_Page}} links of images and the corresponding Wikipedia link for each entity. Text summaries were downloaded using the MediaWiki API\footnote{\url{https://www.mediawiki.org/wiki/API:Main_page}}, while images using the WikiMedia link. The process was accelerated by chunking the data and running multiple sessions. Download errors led to a small loss of entities and the final raw dataset is in Table \ref{tab:mlm_dataset}.

\subsubsection{Geographical cell generation.} \label{geo_cells}
In this stage, entities along with their images and summaries are assigned to geographical cells. To this end, we first exclude all SVG images since they mostly contain maps and signs which are non-descriptive for our tasks. Next, we divide the earth into non-overlapping geographical cells $C$ using the S2 Geometry Library\footnote{\url{https://code.google.com/archive/p/s2-geometry-library/}}. In this library, initial cells are generated by projecting the earth’s surface on a cube with six sides. Based on geocoordinates of the entities, a hierarchical subdivision is applied~\cite{planet16}, where each cell is the node of a quad-tree. Starting at the root nodes, the respective quad-tree is subdivided recursively until all cells contain a maximum of $T_{max}=500$ entities. Finally, in order to have more uniform cells relative to the entities, all generated cells containing less than $T_{min}=20$ entities are removed.

% Section

\section{Availability}
Versions of MLM listed in Table \ref{tab:mlm_dataset} are available for direct download and use. To support findability and sustainability, the MLM dataset is published as an on-line resource at \textbf{ \textit {\url{https://doi.org/10.5281/zenodo.3885753}}}. A separate page with detailed explanations and illustrations is available at \textbf{ \textit {\url{http://cleopatra.ijs.si/goal-mlm/}}} to promote ease-of-use. The project GitHub repository contains the complete source code for the system and generation script is available at \textbf{ \textit {\url{https:/github.com/GOALCLEOPATRA/MLM}}}. Documentation adheres to the standards of \textit{FAIR Data principles}\footnote{\url{http://www.nature.com/articles/sdata201618/}} with all relevant metadata specified to the research community and users. It is freely accessible under the \textit{Creative Commons Attribution 4.0 International license}, making it reusable for almost any purpose. 

\subsection{Updating and Reusability}
MLM is supported by a team of researchers from the University of Bonn, the Leibniz Information Center for Science and Technology, and Jožef Stefan Institute. The resource is already in use for individual projects and as a contribution to the project deliverables of the Marie Skłodowska-Curie CLEOPATRA Innovative Training Network. In addition to the steps above that make the resource available to the wider community, usage of MLM will be promoted to the network of researchers in this project. Awareness among researchers and practitioners in digital humanities will be promoted by demonstrations and presentations at domain-related events. The range of modalities and languages present in the dataset also extends its application to research on multimodal representation learning, multilingual machine learning, information retrieval, location estimation, and the Semantic Web. MLM will be supported and maintained for three years in the first instance. A second release of the dataset is already scheduled and the generation process outlined above is designed to enable rapid scaling.

% Section

\section{Multitask Learning Framework}
In this section, we specify a multitask learning framework that characterises the use case above by performing cross-modal retrieval and location estimation tasks. The section starts with formulations of the tasks evaluated on MLM and concludes with details of the system and methods.

\begin{figure*}
    \centering
    \includegraphics[width=0.6\textwidth, height=0.16\textwidth]{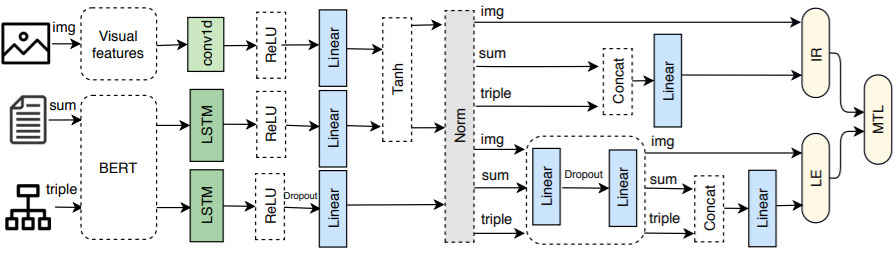}
    \caption{Multitask IR+LE framework.} 
    \vspace{-1em}
    \label{fig:irle_framework}
\end{figure*}

\subsection{Tasks} \label{taskssec}

\subsubsection{Cross-modal Retrieval.}
We build on the approach proposed by Marin et al~\cite{recipe} to perform cross-modal information retrieval on visual $u$ and textual $t$ inputs. Embeddings $v^u$ and $v^t$ are learned as described below. A mapping from the visual and textual embeddings to a common space is learned using a cosine embedding loss, which minimises the distance between $v^u$ and $v^t$ based on cosine similarity and labels $s=\{1, -1\}$
\small
\begin{equation}
L_{cos}(v^u, v^t, l)= \left.
    \begin{cases}
1-cos(v^u_1,v^t_1),& \text{if } s=1\\
max(0,cos(v^u_1,v^t_1) - margin),              & \text{if} s=-1
    \end{cases}
    \right\}
\end{equation}
\normalsize
where $v^u$ and $v^t$ are representations of the visual and textual inputs respectively. At inference time, an image or textual sample is presented with the aim of returning samples for the corresponding modality.

\subsubsection{Location estimation.}
In this architecture, location estimation is regarded as a classification problem. To this end, geo-cells (described in Section \ref{geo_cells}) are regarded as classes. To formulate the location estimation task, we have $q$ geo-cells $y=(y_1,y_2,...y_n)$ for each of which there are $n$ number of entities represented by images, summaries and triples. Each entity is labeled with its corresponding GPS coordinates in (latitude, longitude) pairs. The task is to predict $y$ given $v^u$ and $v^t$ based on a cross-entropy loss:

\small
 \begin{equation}
 L(v,y) = -v[y]+log\left(\sum_jexp(v[j])\right) 
 \end{equation}
\normalsize

\subsubsection{Multitask.} 
The multitask learning framework (see Figure \ref{fig:irle_framework}) trains a combined pipeline to perform cross-modal retrieval and location estimation in conjunction. The system learns over the cross-entropy and cosine embedding losses with a weighted average. In order to account for the difference in magnitude between losses, log variance is calculated over the tasks and converted to log standard deviation $log(\sigma)$. This weight is applied to the losses before computing the mean $\mu$ over the products.  

\subsection{System}
We apply state-of-the-art embedding methods in computer vision and NLP to generate representations of the inputs. Two types of descriptors are utilized to better represent location information of the images when generating Visual embeddings $v^u$. One of the descriptors is taken from a ResNet~\cite{residual} model pre-trained on the Places365 dataset~\cite{zhouplaces}, where the task is to recognize 365 distinct places - examples include beach, stadium, street. The second descriptor is taken from the model \cite{2018geolocation} based on ResNet101~\cite{residual} aimed at predicting the geolocation information of an image. This model is pretrained on a subset of the Yahoo Flickr Creative Commons 100 Million dataset (YFCC100M)~\cite{yfcc100m}. The subset includes around five million geo-tagged images with geolocation labels, and was introduced for the MediaEval Placing Task 2016 (MP-16)~\cite{larson2017benchmarking}. Each descriptor results in a feature vector of size 2048, so eventually we concatenate both vectors and get a vector of size 4096 for each image. To generate textual embeddings $v^t$ for summaries and triples we use BERT~\cite{JacobMing} to extract word vectors for all sentences in text. Since text content of news articles is long, we select maximum 500 characters at a time and extract word vectors from BERT, and average the vectors obtained from each token of the sentence. 

An integrated architecture (see Figure~\ref{fig:mlm_generation}) takes  the visual and text embeddings to perform the two tasks in concurrence. Visual embeddings $v^u$ are fed into a 1D convolutional layer (conv1d) with ReLU activation $max(0,x)$ - and a linear layer with hyperbolic tangent $tanh(u_i^l)$. An LSTM layer replaces the conv1d for text summary $v^s$ and triple class $v^r$ embeddings. Regularisation consists of dropout for textual embeddings and norm with a clamp $min=1*10^{-12}$ for all modalities. Linear layers with dropout receive the outputs of the norm in location estimation. In both tasks, text summary and triple classes are fused with concatenation $v^t=[v^s,v^r]$.

% Section

\section{Experiments}
Experiments assess the performance of the multitask learning system in relation to single task pipelines. We align with benchmark cross-modal retrieval tasks~\cite{recipe} in selecting median rank (medR) as the primary metric. Median rank also takes into account both the performance of the retrieval and the position of the retrieved instance. To compute medR, a subset of 500 instances are randomly sampled from the test set. Then the experiments are repeated 10 times and report the mean results. Next, for each input image or text, items in the corresponding modality are ranked. The objective is to retrieve the sample for the target modality at the highest rank. The retrieved lists are further assessed using recall percentage at top K (R@K). This measures the percentage of queries where the matching item is ranked among the top K returned samples. The formulation of location estimation as a multiclass classification task motivates the selection of Precision (P) (Eq.\ref{subeqn:pre}), Recall (R) (Eq.\ref{subeqn:rec}) and weighted F1 score (Eq.\ref{subeqn:fscore}) as measures of performance for the second task.
\small
\begin{subequations}
\label{eqn:metric}
 \begin{alignat}{6}
   Precision & = \frac{TP }{TP + FP} \label{subeqn:pre} \\[5pt]
   Recall & = \frac{TP }{TP + FN} \label{subeqn:rec} \\[5pt]
   F_1^{weighted} & = \frac{1}{Q^2}\sum_{j=1}^QQ_j\frac{2~ \times P_j \times R_j }{~P_j + R_j} \label{subeqn:fscore}
 \end{alignat}
\end{subequations}
\normalsize
In the above definitions, $TP$, $FP$ and $FN$ stand for True Positive, False Positive and False Negative respectively and $Q$ is the number of geo-cells, and $j$ is the $jth$ instance of geo-cells.

\subsection{System Configuration and Training}
Pipelines for single task and multitask learning are trained for 100 epochs on a single GeForce GTX 1080 Ti GPU, using training/validation sets of 175k/22k and 18k/2k for MLM-irle and MLM-irle-gr respectively. Full details on the dataset splits are available at \textbf{ \textit {\url{http://cleopatra.ijs.si/goal-mlm/}}}. Although we trained for a number of epochs adopted for comparable systems \cite{recipe}, we note that validation loss was still declining for both multitask and single task systems at the end of training. For all tasks, consistent hyperparameter settings are maintained to enable comparative analysis of the results. The batch size is set at 128 with a learning rate of 0.0001 and a probability of 0.1 in dropout layers.

\begin{table*}[h]
\caption{Cross-modal retrieval results for MLM}\label{tab?}
\vspace{-1em}
\begin{tabular}{l l l l l l l l l l}
\hline
                                &                      & \multicolumn{4}{l }{\textbf{Image to Text}} & \multicolumn{4}{l}{\textbf{Text to Image}} \\ \cline{3-10}
                                &                      & \textbf{medR}    & \textbf{R@1}    & \textbf{R@5}    & \textbf{R@10}   & \textbf{medR}   & \textbf{R@1}    & \textbf{R@5}    & \textbf{R@10}   \\ \hline
\multirow{2}{*}{\textbf{MLM-irle}}  & \textbf{Single task} & 43.1    & 0.02   & 0.07   & 0.14   & 32.5   & 0.04   & 0.14   & 0.24   \\ 
                                & \textbf{Multitask}  & 19.9    & 0.05   & 0.20   & 0.34   & 11.9   & 0.11   & 0.34   & 0.47   \\ \hline
\multirow{2}{*}{\textbf{MLM-irle-gr}}    & \textbf{Single task} & 53.6    & 0.01   & 0.06   & 0.13   & 43.2   & 0.03   & 0.10   & 0.18   \\ 
                                & \textbf{Multitask}  & 27.9    & 0.03   & 0.14   & 0.25   & 21.8   & 0.07   & 0.23   & 0.35   \\ \hline
\end{tabular}
\label{table:ir}
\end{table*}

\begin{table}[]
\caption{Location estimation results for MLM}\label{tab?}
\vspace{-1em}
\scalebox{0.9}{
\begin{tabular}{l l l l l l l l}
\hline
                      &                      & \multicolumn{3}{l }{\textbf{Image-based}}       & \multicolumn{3}{l }{\textbf{Text-based}}        \\ \cline{3-8}
                      &                      & \textbf{F1} & \textbf{P} & \textbf{R} & \textbf{F1} & \textbf{P} & \textbf{R} \\ \hline
{\textbf{MLM-}} & \textbf{Single task} & 0.16      & 0.17    & 0.17 & 0.59      & 0.61    & 0.60 \\ 
                      {\textbf{irle}}  & \textbf{Multitask}   & 0.15      & 0.17    & 0.16 & 0.58      & 0.61    & 0.59 \\ \hline
{\textbf{MLM-}}   & \textbf{Single task} & 0.28      & 0.30    & 0.28 & 0.67      & 0.70    & 0.68 \\ 
                    {\textbf{irle-gr}}  & \textbf{Multitask}   & 0.25      & 0.27    & 0.26 & 0.67      & 0.69    & 0.68 \\ \hline
\end{tabular}}
\label{table:le}
\vspace{-4mm}
\end{table}

\subsection{Results}
In this section evaluation results are discussed for multitask and single task pipelines on the MLM-irle and MLM-irle-gr (ie geo-representative) datasets containing 22k and 2k test instances respectively.

\subsubsection{Cross-modal Retrieval}
In Table~\ref{table:ir}, we report medR and Recall@K for image to text (first four columns) and text to image retrieval (the remaining columns).  The multitask system attained a strong improvement on the single task equivalent in the first of these with medR at 19.9. Qualitative checks on samples consisted of extracting a random sample of 500 entities from the test set. Two examples of triple classes where the multitask system scored strongly relative to the single equivalent are “town of the United States” and “Ortsteil” (see Figure \ref{fig:mlm_process}). This corresponds to increases of 13\% on R@5 and 20\% points on R@10. Although R@1 also showed a marked increase, scores for both multitask and single task are at low levels. The multitask system also outperformed strongly on the geo-representative dataset with a 26-point improvement on medR relative to the single task pipeline. Multitask and single task systems attained stronger scores on text to image retrieval for both datasets with double the R@K values and medR at 11.9 - underling the superiority of the multitask approach in cross-modal retrieval. 

\subsubsection{Location Estimation} A summary of performance for multitask and single task systems on location estimation is presented in Table.~\ref{table:le}. Results are broken down by performance on predicting geo-cells using images and text consecutively. As with cross-modal retrieval, both systems return stronger scores when inputs are textual relative to visual. A notable difference with the first task is that performance improves on the geo-representative dataset. This difference is most marked in image-based estimation: the F1 score for the multitask system is 10 points higher on MLM-irle-gr relative to the larger dataset. Scores for the multitask system on both image- and text-based location estimation are closely aligned with the single task system - indicating there are only modest gains from using the former approach on the second task.

% Section

\section{Related Research}
In the first instance, researchers designing multitask systems that learn from multimodal data turn to existing multimodal resources for benchmarking. A reliance on image captioning datasets has led to multitask learning proposals limited to paired image and text samples~\cite{binder2013enhanced, luo2019cross}. Datasets in the medical and bioinformatics areas that have inspired multitask approaches also use these two modalities~\cite{nguyen2019overcoming, KornutaRajan}. Several proposals pair natural language and knowledge graph embeddings for training multitask systems~\cite{luan-etal-2018-multi}. In contrast to our research, tasks in this area are concentrated on applications related to NLP and knowledge graphs - and use text-only data. Multitask learning on multimodal data is an area of research in robotics~\cite{brodeur+al-2018-home-iclr, singh2019embodied} - although the objective of training autonomous agents is distinct from the use cases for MLM. In all of the cited research, text inputs are English-only. Multilingual resources and benchmarks for training multitask architectures are motivated primarily to solve NLP tasks and are correspondingly unimodal~\cite{hu2020xtreme, lin2018multi}. A multitask solution was proposed for the multilingual IKEA dataset with a specific objective in machine translation~\cite{zhou2018visual}. As with previously noted cases, this resource is bimodal and so applies to a smaller subset of constituent tasks relative to MLM.  

Research on information retrieval for multimodal data has an extensive history and we focus here on resources and benchmark tasks where samples are present in multiple languages. The Cross-Language Evaluation Forum (CLEF) evolved into ImageCLEF (from 2003), GeoCLEF (from 2005) and VideoCLEF (from 2008) tracks. Resources in the LifeCLEF tracks of ImageCLEF have included geocoordinates - but the focus is primarily on images and is aimed at researchers in disciplines related to biology~\cite{joly2016lifeclef}. Information retrieval on video content has extended to multilingual scenarios where either queries, content, or both are in several languages~\cite{aytar2008utilizing}. Datasets in this area are mostly composed of comparatively small numbers of samples. Constituent modalities (ie video, audio) vary from those in MLM and place an emphasis on the spatiotemporal context of the inputs~\cite{wang2019vatex}. Video retrieval is also differentiated by an emphasis on domain-specific sub-tasks including shot segmentation~\cite{delezoide2007semanticvox} and detection of objects or scenes~\cite{wang2019vatex}. All of these resources exclude semantic Web elements.

Location estimation methods related to this research breakdown into three main groups~\cite{brejcha2017state}: natural~\cite{baatzlarge,zhengtour}, city-scale~\cite{kimlearned,chen2011city} and global~\cite{kordopatis2016depth,trevisiol2013retrieving}. Several methods in the published research rely on a single modality~\cite{planet16,2018geolocation,middleton2018location} and we focus here on multimodal approaches. MediaEval benchmark placing datasets~\cite{larson2017benchmarking, choi2015placing} include more than five million instances with images, videos and metadata used to estimate capturing locations represented in multimedia. Kordopatis-Zilos et al.~\cite{kordopatis2017geotagging} build on this dataset by combining visual features with a language model based on word frequency in the text for geolocation estimation. In~\cite{BurakEvan} a dataset containing image and text is collected from Wikipedia combined with around 890K satellite images and is used for interpreting satellite images. BreakingNews~\cite{breakingnews} is a resource with 100K instances in the domains of sports, politics, arts, healthcare or local news. The researchers apply a baseline that combines a Convolutional Neural Network and Recurrent Neural Networks for multimodal geolocation estimation. In addition there are other approaches that use geolocation information for image recognition~\cite{feifeiliclassify, chu2019geoaware} and image understanding~\cite{yuseason}. In~\cite{yuseason} the authors leverage season and location context with a probabilistic framework to help improve region recognition in images. In~\cite{chu2019geoaware} the effectiveness of using geolocation on fine-grained recognition is examined.  All the aforementioned methods are based on datasets in English only. All these resources are intended for single task scenarios for the corresponding task. In contrast, this paper provides a multilingual dataset with a great number of modalities. The objective also differs in the focus on systems that perform geolocation estimation as one of a sequence of tasks.

Research on different forms of bias in data - and their impact on machine learning systems - underlines the value of resources that provide balanced coverage of their domains. Representation bias in relation to geographical location is a form of bias that has high relevance to MLM and has been identified in open source data sets used widely to train machine learning systems~\cite{46553}. Representation bias results in poor classification performance on samples for classes that are under-sampled in the data. In another part of the machine learning pipeline, evaluation bias results when the test data fails to match the population of interest~\cite{suresh2019framework}. We are motivated by this research to address representation and evaluation bias in this resource.

% Section

\section{Conclusion and Future Work}
In this research, we introduced the MLM resource for evaluating multitask systems on diverse data. In addition, we specified a set of benchmark tasks that characterise the specific use case of geoparsing and georeferencing information encoded in multiple modalities and languages. A baseline multitask system that makes use of both cross-modal and multimodal fusion was specified and evaluated against single task systems. The paper details a process that generates a resource comprising linked entities. The generation process also enables consistent and rapid scaling of the resource for future releases. Finally we also ensured that the dataset meets FAIR principles and released a geo-representative version of the dataset to serve cultural and academic institutions with a focus on the European Union. We believe that the resource will improve generalisation in multitask learning systems and so streamline the pipelines of applications that learn from materials stored in digital archives. In upcoming research, we will release future versions with data in additional languages, add entries from other data sources, design systems for specific functions, and develop an application that further exploits contextual data. 

\section{Acknowledgments}

The project leading to this publication has received funding from the European Union's Horizon 2020 research and innovation programme under the Marie Skłodowska-Curie grant agreement No 812997.

% Bibliography

\bibliographystyle{plain}
{\footnotesize
\bibliography{bibliography-oa.bib}}

\begin{thebibliography}{10}

\bibitem{alex2015adapting}
Beatrice Alex, Kate Byrne, Claire Grover, and Richard Tobin.
\newblock Adapting the {Edinburgh} geoparser for historical georeferencing.
\newblock {\em International Journal of Humanities and Arts Computing},
  9(1):15--35, 2015.

\bibitem{ali2019keen}
Mehdi Ali, Hajira Jabeen, Charles~Tapley Hoyt, and Jens Lehmann.
\newblock The {KEEN} universe.
\newblock In {\em International Semantic Web Conference}, pages 3--18.
  Springer, 2019.

\bibitem{aytar2008utilizing}
Yusuf Aytar, Mubarak Shah, and Jiebo Luo.
\newblock Utilizing semantic word similarity measures for video retrieval.
\newblock In {\em 2008 IEEE Conference on Computer Vision and Pattern
  Recognition}, pages 1--8. IEEE, 2008.

\bibitem{baatzlarge}
Georges Baatz, Olivier Saurer, Kevin K{\"o}ser, and Marc Pollefeys.
\newblock Large scale visual geo-localization of images in mountainous terrain.
\newblock In {\em European conference on computer vision}, pages 517--530.
  Springer, 2012.

\bibitem{baltruvsaitis2018multimodal}
Tadas Baltru{\v{s}}aitis, Chaitanya Ahuja, and Louis-Philippe Morency.
\newblock Multimodal machine learning: A survey and taxonomy.
\newblock {\em IEEE transactions on pattern analysis and machine intelligence},
  41(2):423--443, 2018.

\bibitem{becker_new_2016}
Tilman Becker, Edward Curry, Anja Jentzsch, and Walter Palmetshofer.
\newblock New {Horizons} for a {Data}-{Driven} {Economy}: {Roadmaps} and
  {Action} {Plans} for {Technology}, {Businesses}, {Policy}, and {Society}.
\newblock In José~María Cavanillas, Edward Curry, and Wolfgang Wahlster,
  editors, {\em New {Horizons} for a {Data}-{Driven} {Economy}: {A} {Roadmap}
  for {Usage} and {Exploitation} of {Big} {Data} in {Europe}}, pages 277--291.
  Springer International Publishing, Cham, 2016.

\bibitem{bengio_learning_2009}
Yoshua Bengio.
\newblock Learning {Deep} {Architectures} for {AI}.
\newblock {\em Foundations and Trends in Machine Learning}, 2:71, 2009.

\bibitem{binder2013enhanced}
Alexander Binder, Wojciech Samek, Klaus-Robert M{\"u}ller, and Motoaki
  Kawanabe.
\newblock Enhanced representation and multi-task learning for image annotation.
\newblock {\em Computer Vision and Image Understanding}, 117(5):466--478, 2013.

\bibitem{blanc2018semi}
Nicolas Blanc, Timoth{\'e}e Produit, and Jens Ingensand.
\newblock A semi-automatic tool to georeference historical landscape images.
\newblock Technical report, PeerJ Preprints, 2018.

\bibitem{brejcha2017state}
Jan Brejcha and Martin {\v{C}}ad{\'\i}k.
\newblock State-of-the-art in visual geo-localization.
\newblock {\em Pattern Analysis and Applications}, 20(3):613--637, 2017.

\bibitem{brodeur+al-2018-home-iclr}
Simon Brodeur, Ethan Perez, Ankesh Anand, Florian Golemo, Luca Celotti, Florian
  Strub, Jean Rouat, Hugo Larochelle, and Aaron~C. Courville.
\newblock Home: a household multimodal environment.
\newblock In {\em 6th International Conference on Learning Representations,
  {ICLR} 2018, Vancouver, BC, Canada, April 30 - May 3, 2018, Workshop Track
  Proceedings}. OpenReview.net, 2018.

\bibitem{caputo2014imageclef}
Barbara Caputo, Henning M{\"u}ller, Jesus Martinez-Gomez, Mauricio Villegas,
  Burak Acar, Novi Patricia, Neda Marvasti, Suzan {\"U}sk{\"u}darl{\i}, Roberto
  Paredes, Miguel Cazorla, et~al.
\newblock Imageclef 2014: Overview and analysis of the results.
\newblock In {\em International Conference of the Cross-Language Evaluation
  Forum for European Languages}, pages 192--211. Springer, 2014.

\bibitem{caruana1997multitask}
Rich Caruana.
\newblock Multitask learning.
\newblock {\em Machine learning}, 28(1):41--75, 1997.

\bibitem{chen2011city}
David~M Chen, Georges Baatz, Kevin K{\"o}ser, Sam~S Tsai, Ramakrishna
  Vedantham, Timo Pylv{\"a}n{\"a}inen, Kimmo Roimela, Xin Chen, Jeff Bach, Marc
  Pollefeys, et~al.
\newblock City-scale landmark identification on mobile devices.
\newblock In {\em CVPR 2011}, pages 737--744. IEEE, 2011.

\bibitem{choi2015placing}
Jaeyoung Choi, Claudia Hauff, Olivier Van~Laere, and Bart Thomee.
\newblock The placing task at {MediaEval} 2015.
\newblock In {\em MediaEval 2015, Wurzen, Germany, 14-15 September 2015; Ceur
  Workshop Proceedings 1436, 2015}. CEUR, 2015.

\bibitem{choi_intra-modal_2019}
Jaeyoung Choi, Martha Larson, Gerald Friedland, and Alan Hanjalic.
\newblock From {Intra}-{Modal} to {Inter}-{Modal} {Space}: {Multi}-task
  {Learning} of {Shared} {Representations} for {Cross}-{Modal} {Retrieval}.
\newblock In {\em 2019 {IEEE} {Fifth} {International} {Conference} on
  {Multimedia} {Big} {Data} ({BigMM})}, pages 1--10, Singapore, Singapore,
  September 2019. IEEE.

\bibitem{chu2019geoaware}
Grace Chu, Brian Potetz, Weijun Wang, Andrew Howard, Yang Song, Fernando
  Brucher, Thomas Leung, and Hartwig Adam.
\newblock Geo-aware networks for fine-grained recognition.
\newblock In {\em Proceedings of the IEEE International Conference on Computer
  Vision Workshops}, pages 0--0, 2019.

\bibitem{delezoide2007semanticvox}
Bertrand Delezoide and Herv{\'e} Le~Borgne.
\newblock Semanticvox: A multilingual video search engine.
\newblock In {\em Proceedings of the 6th ACM international conference on Image
  and video retrieval}, pages 81--84, 2007.

\bibitem{JacobMing}
Jacob Devlin, Ming-Wei Chang, Kenton Lee, and Kristina Toutanova.
\newblock {BERT}: Pre-training of deep bidirectional transformers for language
  understanding.
\newblock In {\em Proceedings of the 2019 Conference of the North {A}merican
  Chapter of the Association for Computational Linguistics: Human Language
  Technologies, Volume 1 (Long and Short Papers)}, pages 4171--4186,
  Minneapolis, Minnesota, June 2019. Association for Computational Linguistics.

\bibitem{harrach2019interactive}
Mouna Harrach, Alexandre Devaux, and Mathieu Br{\'e}dif.
\newblock Interactive image geolocalization in an immersive web application.
\newblock {\em International Archives of the Photogrammetry, Remote Sensing \&
  Spatial Information Sciences}, 2019.

\bibitem{residual}
Kaiming He, Xiangyu Zhang, Shaoqing Ren, and Jian Sun.
\newblock Identity mappings in deep residual networks.
\newblock In {\em European conference on computer vision}, pages 630--645.
  Springer, 2016.

\bibitem{hu2020xtreme}
Junjie Hu, Sebastian Ruder, Aditya Siddhant, Graham Neubig, Orhan Firat, and
  Melvin Johnson.
\newblock Xtreme: A massively multilingual multi-task benchmark for evaluating
  cross-lingual generalization.
\newblock {\em arXiv preprint arXiv:2003.11080}, 2020.

\bibitem{joly2016lifeclef}
Alexis Joly, Herv{\'e} Go{\"e}au, Herv{\'e} Glotin, Concetto Spampinato, Pierre
  Bonnet, Willem-Pier Vellinga, Julien Champ, Robert Planqu{\'e}, Simone
  Palazzo, and Henning M{\"u}ller.
\newblock Lifeclef 2016: multimedia life species identification challenges.
\newblock In {\em International Conference of the Cross-Language Evaluation
  Forum for European Languages}, pages 286--310. Springer, 2016.

\bibitem{kimlearned}
Hyo~Jin Kim, Enrique Dunn, and Jan-Michael Frahm.
\newblock Learned contextual feature reweighting for image geo-localization.
\newblock In {\em 2017 IEEE Conference on Computer Vision and Pattern
  Recognition (CVPR)}, pages 3251--3260. IEEE, 2017.

\bibitem{kordopatis2017geotagging}
Giorgos Kordopatis-Zilos, Symeon Papadopoulos, and Ioannis Kompatsiaris.
\newblock Geotagging text content with language models and feature mining.
\newblock {\em Proceedings of the IEEE}, 105(10):1971--1986, 2017.

\bibitem{kordopatis2016depth}
Giorgos Kordopatis-Zilos, Symeon Papadopoulos, and Yiannis Kompatsiaris.
\newblock In-depth exploration of geotagging performance using sampling
  strategies on yfcc100m.
\newblock In {\em Proceedings of the 2016 ACM Workshop on Multimedia COMMONS},
  pages 3--10, 2016.

\bibitem{KornutaRajan}
Tomasz Kornuta, Deepta Rajan, Chaitanya Shivade, Alexis Asseman, and Ahmet~S.
  Ozcan.
\newblock Leveraging medical visual question answering with supporting facts.
\newblock In Linda Cappellato, Nicola Ferro, David~E. Losada, and Henning
  M{\"{u}}ller, editors, {\em Working Notes of {CLEF} 2019 - Conference and
  Labs of the Evaluation Forum, Lugano, Switzerland, September 9-12, 2019},
  volume 2380 of {\em {CEUR} Workshop Proceedings}. CEUR-WS.org, 2019.

\bibitem{kuga2017multi}
Ryohei Kuga, Asako Kanezaki, Masaki Samejima, Yusuke Sugano, and Yasuyuki
  Matsushita.
\newblock Multi-task learning using multi-modal encoder-decoder networks with
  shared skip connections.
\newblock In {\em Proceedings of the IEEE International Conference on Computer
  Vision Workshops}, pages 403--411, 2017.

\bibitem{larson2017benchmarking}
Martha Larson, Mohammad Soleymani, Guillaume Gravier, Bogdan Ionescu, and
  Gareth~JF Jones.
\newblock The benchmarking initiative for multimedia evaluation: Mediaeval
  2016.
\newblock {\em IEEE MultiMedia}, 24(1):93--96, 2017.

\bibitem{li2018vqa}
Qing Li, Qingyi Tao, Shafiq Joty, Jianfei Cai, and Jiebo Luo.
\newblock Vqa-e: Explaining, elaborating, and enhancing your answers for visual
  questions.
\newblock In {\em Proceedings of the European Conference on Computer Vision
  (ECCV)}, pages 552--567, 2018.

\bibitem{lin2018multi}
Ying Lin, Shengqi Yang, Veselin Stoyanov, and Heng Ji.
\newblock A multi-lingual multi-task architecture for low-resource sequence
  labeling.
\newblock In {\em Proceedings of the 56th Annual Meeting of the Association for
  Computational Linguistics (Volume 1: Long Papers)}, pages 799--809, 2018.

\bibitem{long2017learning}
Mingsheng Long, Zhangjie Cao, Jianmin Wang, and S~Yu Philip.
\newblock Learning multiple tasks with multilinear relationship networks.
\newblock In {\em Advances in neural information processing systems}, pages
  1594--1603, 2017.

\bibitem{luan-etal-2018-multi}
Yi~Luan, Luheng He, Mari Ostendorf, and Hannaneh Hajishirzi.
\newblock Multi-task identification of entities, relations, and coreference for
  scientific knowledge graph construction.
\newblock In {\em Proceedings of the 2018 Conference on Empirical Methods in
  Natural Language Processing}, pages 3219--3232, Brussels, Belgium, 2018.
  Association for Computational Linguistics.

\bibitem{luo2019cross}
Junyu Luo, Ying Shen, Xiang Ao, Zhou Zhao, and Min Yang.
\newblock Cross-modal image-text retrieval with multitask learning.
\newblock In {\em Proceedings of the 28th ACM International Conference on
  Information and Knowledge Management}, pages 2309--2312, 2019.

\bibitem{recipe}
Javier Marin, Aritro Biswas, Ferda Ofli, Nicholas Hynes, Amaia Salvador, Yusuf
  Aytar, Ingmar Weber, and Antonio Torralba.
\newblock Recipe1m+: A dataset for learning cross-modal embeddings for cooking
  recipes and food images.
\newblock {\em IEEE transactions on pattern analysis and machine intelligence},
  2019.

\bibitem{middleton2018location}
Stuart~E Middleton, Giorgos Kordopatis-Zilos, Symeon Papadopoulos, and Yiannis
  Kompatsiaris.
\newblock Location extraction from social media: Geoparsing, location
  disambiguation, and geotagging.
\newblock {\em ACM Transactions on Information Systems (TOIS)}, 36(4):1--27,
  2018.

\bibitem{moncla2017automated}
Ludovic Moncla, Mauro Gaio, Thierry Joliveau, and Yves-Fran{\c{c}}ois~Le Lay.
\newblock Automated geoparsing of {Paris} street names in 19th century novels.
\newblock In {\em Proceedings of the 1st ACM SIGSPATIAL Workshop on Geospatial
  Humanities}, pages 1--8, 2017.

\bibitem{2018geolocation}
Eric Muller-Budack, Kader Pustu-Iren, and Ralph Ewerth.
\newblock Geolocation estimation of photos using a hierarchical model and scene
  classification.
\newblock In {\em Proceedings of the European Conference on Computer Vision
  (ECCV)}, pages 563--579, 2018.

\bibitem{nguyen2019overcoming}
Binh~D Nguyen, Thanh-Toan Do, Binh~X Nguyen, Tuong Do, Erman Tjiputra, and
  Quang~D Tran.
\newblock Overcoming data limitation in medical visual question answering.
\newblock In {\em International Conference on Medical Image Computing and
  Computer-Assisted Intervention}, pages 522--530. Springer, 2019.

\bibitem{pauwels2012multimodal}
Luc Pauwels.
\newblock A multimodal framework for analyzing websites as cultural
  expressions.
\newblock {\em Journal of Computer-Mediated Communication}, 17(3):247--265,
  2012.

\bibitem{prado_m}
Miguel~De Prado, Jing Su, Rabia Saeed, Lorenzo Keller, Noelia Vallez, Andrew
  Anderson, David Gregg, Luca Benini, Tim Llewellynn, Nabil Ouerhani, Rozenn
  Dahyot, and Nuria Pazos.
\newblock Bonseyes ai pipeline—bringing ai to you: End-to-end integration of
  data, algorithms, and deployment tools.
\newblock {\em ACM Trans. Internet Things}, 1(4), August 2020.

\bibitem{breakingnews}
Arnau Ramisa, Fei Yan, Francesc Moreno-Noguer, and Krystian Mikolajczyk.
\newblock Breakingnews: Article annotation by image and text processing.
\newblock {\em IEEE transactions on pattern analysis and machine intelligence},
  40(5):1072--1085, 2017.

\bibitem{46553}
Shreya Shankar, Yoni Halpern, Eric Breck, James Atwood, Jimbo Wilson, and
  D.~Sculley.
\newblock No classification without representation: Assessing geodiversity
  issues in open data sets for the developing world.
\newblock In {\em NIPS 2017 workshop: Machine Learning for the Developing
  World}, 2017.

\bibitem{singh2019embodied}
Devendra Singh~Chaplot, Lisa Lee, Ruslan Salakhutdinov, Devi Parikh, and Dhruv
  Batra.
\newblock Embodied multimodal multitask learning.
\newblock {\em arXiv preprint arXiv:1902.01385}, 2019.

\bibitem{suresh2019framework}
Harini Suresh and John~V Guttag.
\newblock A framework for understanding unintended consequences of machine
  learning.
\newblock {\em arXiv preprint arXiv:1901.10002}, 2019.

\bibitem{feifeiliclassify}
Kevin Tang, Manohar Paluri, Li~Fei-Fei, Rob Fergus, and Lubomir Bourdev.
\newblock Improving image classification with location context.
\newblock In {\em Proceedings of the IEEE international conference on computer
  vision}, pages 1008--1016, 2015.

\bibitem{yfcc100m}
Bart Thomee, David~A Shamma, Gerald Friedland, Benjamin Elizalde, Karl Ni,
  Douglas Poland, Damian Borth, and Li-Jia Li.
\newblock Yfcc100m: The new data in multimedia research.
\newblock {\em Communications of the ACM}, 59(2):64--73, 2016.

\bibitem{trevisiol2013retrieving}
Michele Trevisiol, Herv{\'e} J{\'e}gou, Jonathan Delhumeau, and Guillaume
  Gravier.
\newblock Retrieving geo-location of videos with a divide \& conquer
  hierarchical multimodal approach.
\newblock In {\em Proceedings of the 3rd ACM conference on International
  conference on multimedia retrieval}, pages 1--8, 2013.

\bibitem{tsikrika2011overview}
Theodora Tsikrika, Adrian Popescu, and Jana Kludas.
\newblock Overview of the wikipedia image retrieval task at imageclef 2011.
\newblock In {\em CLEF (Notebook Papers/Labs/Workshop)}, volume~4, page~5,
  2011.

\bibitem{BurakEvan}
Burak Uzkent, Evan Sheehan, Chenlin Meng, Zhongyi Tang, Marshall Burke, David~B
  Lobell, and Stefano Ermon.
\newblock Learning to interpret satellite images using wikipedia.
\newblock In {\em IJCAI}, pages 3620--3626, 2019.

\bibitem{wang2019vatex}
Xin Wang, Jiawei Wu, Junkun Chen, Lei Li, Yuan-Fang Wang, and William~Yang
  Wang.
\newblock Vatex: A large-scale, high-quality multilingual dataset for
  video-and-language research.
\newblock In {\em Proceedings of the IEEE International Conference on Computer
  Vision}, pages 4581--4591, 2019.

\bibitem{planet16}
Tobias Weyand, Ilya Kostrikov, and James Philbin.
\newblock Planet-photo geolocation with convolutional neural networks.
\newblock In {\em European Conference on Computer Vision}, pages 37--55.
  Springer, 2016.

\bibitem{xue-wen_chen_big_2014}
{Xue-Wen Chen} and {Xiaotong Lin}.
\newblock Big {Data} {Deep} {Learning}: {Challenges} and {Perspectives}.
\newblock {\em IEEE Access}, 2:514--525, 2014.

\bibitem{yuseason}
Jie Yu and Jiebo Luo.
\newblock Leveraging probabilistic season and location context models for scene
  understanding.
\newblock In {\em Proceedings of the 2008 international conference on
  Content-based image and video retrieval}, pages 169--178, 2008.

\bibitem{zhang_survey_2018}
Yu~Zhang and Qiang Yang.
\newblock A {Survey} on {Multi}-{Task} {Learning}.
\newblock {\em arXiv:1707.08114 [cs]}, July 2018.
\newblock arXiv: 1707.08114.

\bibitem{zhengtour}
Yan-Tao Zheng, Ming Zhao, Yang Song, Hartwig Adam, Ulrich Buddemeier,
  Alessandro Bissacco, Fernando Brucher, Tat-Seng Chua, and Hartmut Neven.
\newblock Tour the world: building a web-scale landmark recognition engine.
\newblock In {\em 2009 IEEE Conference on Computer Vision and Pattern
  Recognition}, pages 1085--1092. IEEE, 2009.

\bibitem{zhouplaces}
Bolei Zhou, Agata Lapedriza, Aditya Khosla, Aude Oliva, and Antonio Torralba.
\newblock Places: A 10 million image database for scene recognition.
\newblock {\em IEEE transactions on pattern analysis and machine intelligence},
  40(6):1452--1464, 2017.

\bibitem{zhou2018visual}
Mingyang Zhou, Runxiang Cheng, Yong~Jae Lee, and Zhou Yu.
\newblock A visual attention grounding neural model for multimodal machine
  translation.
\newblock {\em arXiv preprint arXiv:1808.08266}, 2018.

\end{thebibliography}

\end{document}